\documentclass[letterpaper, 10 pt, conference]{ieeeconf}  

\IEEEoverridecommandlockouts                              

\usepackage{mathptmx} 
\usepackage{times} 
\usepackage{amsmath} 
\usepackage{amssymb}  
\usepackage{booktabs}
\usepackage{float}
\usepackage{capt-of}
\usepackage{xcolor}
\usepackage{xspace}
\usepackage{graphicx}
\usepackage{multirow,multicol}
\usepackage{cuted}
\usepackage[%
backend=biber
,style=ieee
,isbn=false
,url=false
,doi=false
,eprint=false
]{biblatex}
\usepackage{censor}
\usepackage{pifont}
\usepackage{makecell}
\usepackage[hidelinks]{hyperref}

\def\BibTeX{{\rm B\kern-.05em{\sc i\kern-.025em b}\kern-.08em
    T\kern-.1667em\lower.7ex\hbox{E}\kern-.125emX}}
\title{\LARGE \bf
SILICA: Repurposing Diffusion Priors for Joint Glass Segmentation and Depth Estimation}

\author{{Tarun R}$^{1}$, \quad {Anuj Verma}$^{1}$, \quad {Laksh Nanwani}$^{1}$, \quad {Sourav Garg}$^{1}$, \quad {K. Madhava Krishna}$^{1}$
\thanks{$^{1}$ {Robotics Research Center (RRC), IIIT Hyderabad}}
\thanks{Project Page - \href{https://silica-mirage.github.io/}{https://silica-mirage.github.io/}}
\thanks{Code - \href{https://github.com/rtarun1/Silica}{https://github.com/rtarun1/Silica}}
\thanks{Dataset - \href{https://huggingface.co/datasets/rtarun1/mirage18k}{Mirage18k}}
}

\begin{document}

\maketitle
\thispagestyle{empty}
\pagestyle{empty}

\begin{strip}
\centering
\vspace{-2cm}
\includegraphics[width=\linewidth]{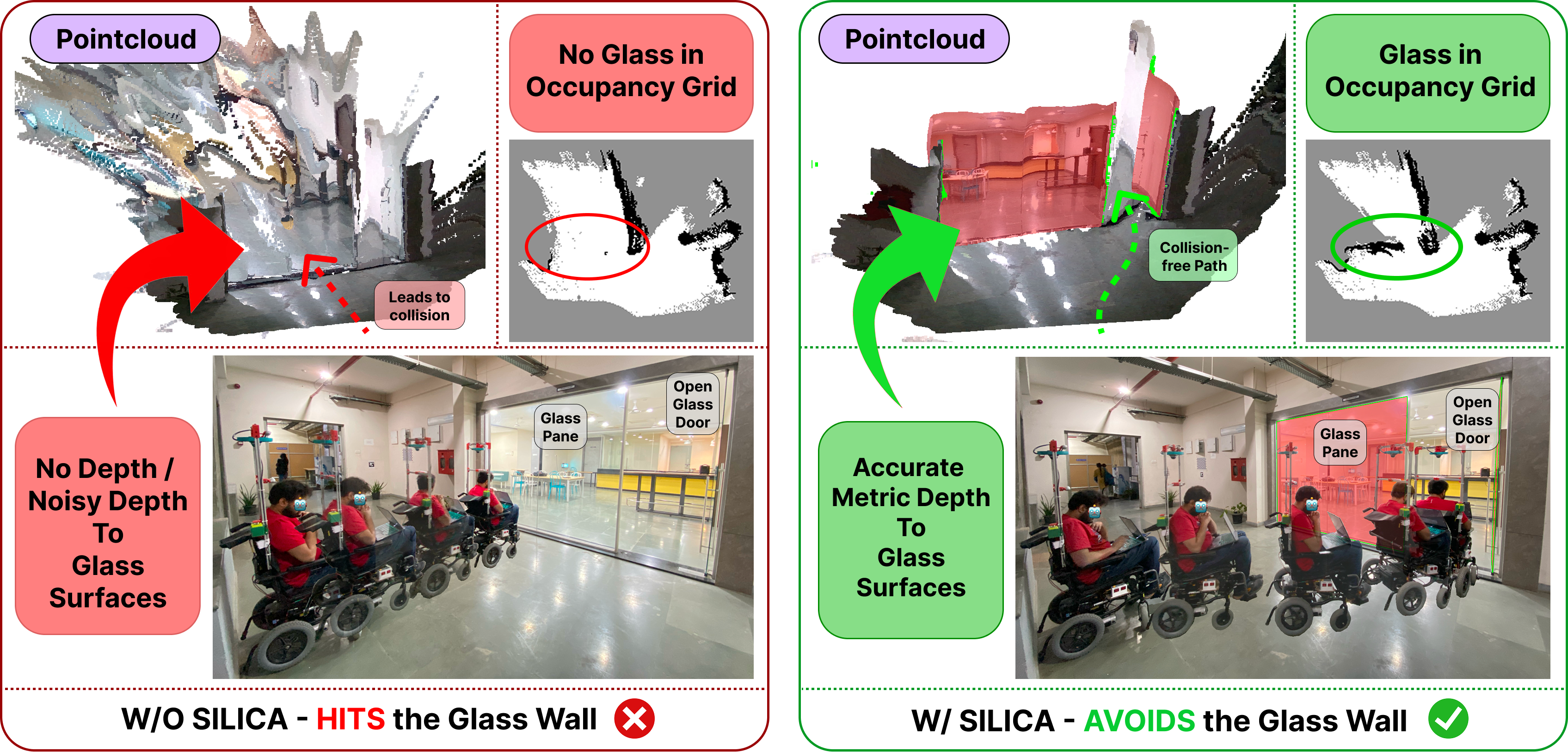}
\captionof{figure}{(Left) Conventional depth cameras and LiDAR sensors do not provide correct depth to glass, leading to misinformed mapping and navigation, potentially allowing robots to collide with transparent structures. (Right) SILICA overcomes this inherent limitation by jointly segmenting glass in an image and predicting accurate metric depth, allowing our robot to identify the correct open door and plan a safe path.}

\label{fig:fig_teaser}
\end{strip}

\begin{abstract}
Standard depth sensors systematically fail on transparent surfaces, creating corrupted 3D maps and severe navigation hazards. While specialized hardware sensors can detect glass, they lack modularity and have extensive hardware dependencies. Consequently, learning-based monocular depth estimation has emerged as a compelling alternative. However, domain-specific glass-aware monocular depth estimators struggle with unfamiliar indoor layouts; restricted by the severe scarcity of real-world glass depth annotations, they fail to generalize zero-shot to new settings. This motivates us to explore whether the extensive priors of text-to-image diffusion models can enable generalizable perception of transparent surfaces. We introduce SILICA, a unified pipeline leveraging these priors to jointly predict glass segmentation and glass-aware depth. This mutual information exchange establishes a robust spatial hierarchy, entirely eliminating the need for paired real-world glass depth annotations. Subsequently, we use the predicted segmentation mask to explicitly filter incorrect glass depth points from standard sensors, recovering accurate metric glass depth for downstream 3D mapping and autonomous collision avoidance. Supported by our novel Mirage 18k dataset, extensive experiments demonstrate that SILICA achieves remarkable zero-shot transfer across diverse, unseen environments, outperforming state-of-the-art models by almost 20\% and setting a new benchmark for transparent surface perception.
\end{abstract}
\section{INTRODUCTION}

Perceiving transparent structures, such as glass panes, doors, and partitions, remains a fundamental yet highly challenging task in robotic perception. While recovering accurate 3D scene geometry is essential for safe navigation and SLAM, standard depth sensors like LiDAR, RGB-D, and stereo cameras fail on these surfaces due to specular reflection, refraction, and visual ambiguity. Although specialized acoustic sensors offer a hardware solution, their sparse outputs and strict calibration dependencies make them inherently unreliable. As a result, learning-based monocular depth estimation has become a prominent software alternative. Recent efforts have introduced specialized, glass-aware networks \cite{GWDepth} trained in a supervised manner on paired collections of RGB and metric depth maps. Unfortunately, it is extremely challenging to collect real-world glass annotations restricts these models to narrow domains, causing them to struggle when generalizing to unfamiliar scenes.

Early learning based methods attempted to solve this by predicting 2D boundary segmentation masks \cite{eblnet, TransLab_Trans10k, GlassSemNet_GSDS, GDD_DGNet}, but these lack the explicit 3D geometric context required for navigation. More recently, the advent of foundational monocular depth estimators \cite{depth_anything_v2} has brought about a leap in general scene understanding. Still, these foundational depth estimators tend to struggle when presented with a complex transparent structure. In our experiments, these methods consistently fail when prominent features are visible behind the glass, predicting the depth of the background rather than the physical surface of the glass itself.

Our core intuition is that accurately perceiving transparent structures requires resolving the severe foreground-background ambiguities inherent to these surfaces. We hypothesize that jointly learning glass segmentation and monocular depth estimation provides a mutually beneficial solution. Under this hypothesis, we posit that an inferred glass mask would act as an explicit spatial feature to guide surface depth, while geometric depth context would help the segmentation network isolate objects in front of the glass from the visible background. However, attempting this dual-task framework exposes a severe domain disparity of datasets, where available glass segmentation datasets consist entirely of real-world imagery, whereas dense, artifact-free depth supervision necessitates synthetic data \cite{hypersim}.

To bridge this domain gap and achieve generalizable joint perception, we introduce SILICA: a unified pipeline leveraging the rich priors of text-to-image diffusion models \cite{stable_diffusion}. To manage this dual-task framework, we exploit the network's inherent CLIP \cite{clip} conditioning by injecting task-specific text embeddings directly into the U-Net's pre-trained cross-attention layers. This explicit routing facilitates robust cross-modal mutual information exchange, allowing real-world segmentation features to inherently guide synthetic affine-invariant depth estimation, and vice versa. This bypasses the need for paired real-world glass depth annotations. To satisfy the latency constraints of robotics, we collapse the iterative generative process into an efficient, single-step deterministic latent regression model. During real-world deployment, SILICA effectively utilizes its predicted mask to filter incorrect readings on transparent surfaces from standard depth sensors. The remaining reliable, non-glass points are then used to align our affine-invariant predictions into accurate glass-aware metric depth. As a result, SILICA exhibits exceptional zero-shot generalization and attains state-of-the-art performance in complex indoor environments.

Our main contributions are summarized as follows:
\begin{itemize}
    \item \textbf{The SILICA Pipeline:} An efficient, single-step joint perception framework that exploits text-to-image diffusion priors. By leveraging CLIP-conditioned cross-attention, it drives robust mutual information transfer between real-world segmentation and synthetic depth pathways, entirely bypassing the need for paired real-world glass depth annotations. Enabling SILICA to set new benchmark for transparent surface perception. In completely unseen environments, our approach demonstrates exceptional zero-shot capabilities, outperforming existing state-of-the-art models by nearly 20\% in  specific cases.
    \item \textbf{The Mirage 18k Dataset:} A novel, dataset comprising 18,353 complex indoor RGB images paired with binary glass segmentation masks, alongside a evaluation subset of 2,406 samples with corresponding ground-truth metric depth for transparent surfaces.
\end{itemize}

\section{RELATED WORKS}

\subsection{Transparent Surface Segmentation}
Traditional approaches to perceiving transparent surfaces often frame the problem as a 2D segmentation task. Boundary-centric models \cite{TransLab_Trans10k,eblnet} rely on the high visual contrast of glass edges, but fail on frame-less doors or complex glass panes where edge cues are scarce. Conversely, context-driven networks \cite{GDD_DGNet, GlassSemNet_GSDS} exploit high-level semantic co-occurrences, but also fail when presented with complex, cluttered backgrounds visible through the glass. Even recent, massive state-of-the-art models like SAM3~\cite{sam3} struggle in the presence of salient features behind the glass surface.

\subsection{Transparent Glass Datasets}
Existing glass segmentation datasets \cite{TransLab_Trans10k, GDD_DGNet, GlassSemNet_GSDS} offer visual diversity but strictly provide 2D segmentation masks, lacking the metric depth annotation. While manipulation-focused 3D datasets \cite{ClearGrasp, TransCG, KeyPose} provide dense depth, they are restricted to small tabletop objects. Recent attempts to capture real-world planar glass depth, such as GWDepth~\cite{GWDepth}, are severely limited in scale and diversity (comprising only $\sim$1,200 samples). This lack of paired, diverse, real-world 3D glass data fundamentally motivates SILICA's training strategy, which bridges real-world segmentation masks with synthetic depth annotations.

\subsection{Monocular Depth Estimation Methods}
Estimating depth ``in the wild" refers to methods that are successful across a wide range of (possibly unfamiliar) settings. While early methods predicted relative depth, its limited geometric utility drove a shift toward affine-invariant depth, which recovers structure up to an unknown global scale and offset. Recently, foundational models \cite{depth_anything_v2, depthanything} have achieved exceptional affine-invariant zero-shot performance by training massive discriminative architectures on large-scale synthetic and pseudo-labeled datasets. However, despite this massive data coverage, these models still fundamentally fail to resolve depth for transparent surfaces in complex settings, incorrectly predicting the visible background instead.

\subsection{Diffusion-based Monocular Depth Estimation}
Rather than training massive networks from scratch, state-of-the-art methods repurpose the encyclopedic spatial priors of text-to-image Latent Diffusion Models (LDMs) like Stable Diffusion. Operating in a compressed latent space, LDMs implicitly capture complex structural hierarchies from internet-scale data. Breakthroughs like Marigold~\cite{marigold} and GeoWizard~\cite{geowizard} successfully adapt these generative priors for affine-invariant depth, achieving remarkable zero-shot generalization while enabling highly resource-efficient fine-tuning. However, unlike these previous methods, which systematically discard the multi-modal conditioning pathway, we leverage the model's full architectural capacity. We propose a targeted fine-tuning protocol that explicitly re-engages this conditional alignment, achieving robust, in-the-wild performance for our joint task (as demonstrated in Fig.~\ref{fig:qualitative}).

\section{METHODOLOGY}

\begin{figure*}[h]
    \centering
    \includegraphics[width=\textwidth]{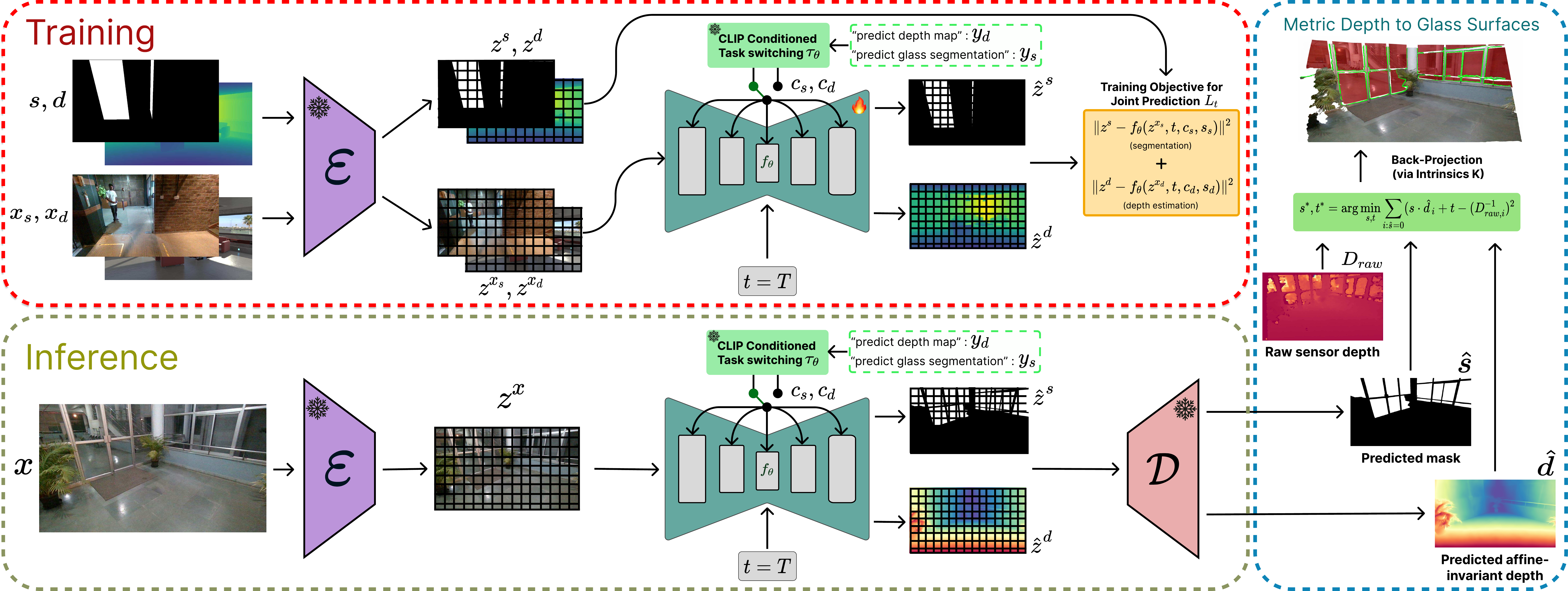}
    \caption{\textbf{SILICA Pipeline Overview.} Our framework employs a shared U-Net architecture for joint glass segmentation and monocular depth estimation $\hat{z}^s, \hat{z}^d$. During Training (top), we employ a fixed timestep $t=T$ joint latent regression loss $L_T$ using CLIP-conditioning $c_s, c_d$ for task switching, where the latent U-Net learns to directly predict the annotation latents $z^s, z^d$ from input latents $z^{x_s}, z^{x_d}$. During Inference (bottom), the model simultaneously predicts task-specific latents, which are decoded and processed through a least-squares optimization to recover aligned metric depth for transparent surfaces.}
    \label{fig:pipeline}
\end{figure*}

The primary objective of the SILICA is to initially take a single input image $x$ and jointly predict a glass segmentation mask $\hat{s}$ alongside a glass-aware, affine-invariant depth map $\hat{d}$. Subsequently, these joint predictions are utilized in a least-squares optimization framework to align $\hat{d}$ with noisy metric depth $D_{raw}$ samples obtained from a depth sensor. Crucially, the predicted segmentation mask $\hat{s}$ is used to explicitly filter out incorrect sensor depth measurements on transparent surfaces during this alignment, ultimately producing an accurate, 3D metric depth map with glass structures outlined for downstream tasks. An overview of the complete SILICA pipeline is illustrated in Fig.~\ref{fig:pipeline}.

\subsection{Latent Regression}
\label{sec:latent_regression}

For diffusion training, there are two basic parameterizations of the diffusion loss function $L_t$:
\begin{enumerate}
    \item $\epsilon${-prediction:} The model $f_\theta^\epsilon$ learns to predict the added noise $\epsilon$ \cite{ddpm}:
    $$L_t^{\epsilon} = \left\| \epsilon - f_\theta^{\epsilon}\!\left({z}_t^{y}, {z}^{x}, t\right) \right\|^2$$
    \item $x_0${-prediction:} The model $f_\theta^{{z}}$ learns to directly predict the clean target sample ${z}^{y}$ \cite{ddpm}:
    $$L_t^{{z}} = \left\| {z}^{y} - f_\theta^{{z}}\!\left({z}_t^{y}, {z}^{x}, t\right) \right\|^2$$
\end{enumerate}

Previous literature \cite{Benny_2022_CVPR, DBLP:journals/corr/abs-2202-00512} indicates that while $\epsilon$-prediction relies on scaling coefficients that severely magnify output variance at high noise levels, $x_0$-prediction directly estimates the clean target sample, thereby avoiding this instability and yielding much steadier predictions. Furthermore, fixing the timestep to $t = T$ eliminates the possibility of error propagation during the multi-step denoising process \cite{lotus}. However, due to the stochastic nature of diffusion models, feeding random noise to the input latent still yields slightly different outputs for different random seeds, which was empirically shown by prior work~\cite{lotus}.

For our robotics perception task, it is essential to guarantee consistent outputs regardless of external factors. Hence, we opt to remove the Gaussian noise addition ${z}^{y}_t$ entirely. Instead, we directly feed the image latent ${z}^{x}$ into the U-Net at the fixed timestep $t = T$:

\begin{equation}
L_t = \left\|{{z}}^{y} - f_\theta({z}^{x}, t) \right\|^2.
\label{eq:x_0_without_noise}
\end{equation}
This deterministic formulation ensures consistent predictions, and allows the pipeline to train and infer more efficiently, even with limited data.

\subsection{Joint Modeling of Glass Segmentation and Depth Estimation}
\label{sec:clip_c_joint_prediction}

\textit{Sanity check for using the same VAE and Decoder for joint task:} While prior work~\cite{marigold} demonstrated that the frozen Stable Diffusion Variational Autoencoder (VAE) $\mathcal{E}$ can accurately reconstruct affine-invariant depth maps with low error(even though being pretrained on RGB), a similar re-use for segmentation or joint modeling of segmentation and depth estimation may not be assumed. Thus, we conduct an initial sanity check to verify if the frozen VAE $\mathcal{E}$ and Decoder $\mathcal{D}$ can accurately reconstruct binary glass segmentation. For this, we normalize the single-channel ground truth mask to the operational range of the VAE, i.e., $[-1, 1]$, and replicate it across three channels to simulate an RGB input tensor $s$. We then pass it through the frozen VAE Encoder to obtain its latent representation $z^s =\mathcal{E}(s)$, and to recover latent binary segmentation $z^s$, it is immediately passed through the frozen Decoder $\hat{z}^s = \mathcal{D}(z^s)$. Subsequently, averaging the three output channels yields an exceptional reconstruction mean Intersection over Union (mIoU) of 99.97\%, demonstrating that latent compression and recovery can be done with minimal error. This enables us to use the same VAE and Decoder for training multi-modal data without using a domain-specific Encoder Decoder. 

We not only have to deal with multi-modal data but also sim and real data mix. Due to the limitations of datasets containing both glass segmentation masks and accurate depth-to-glass annotations, our training configuration inevitably introduces a significant domain disparity: our segmentation pairs $({s}, {x}_s)$ consist entirely of real-world samples, while our depth samples $({d}, {x}_d)$ are entirely synthetic. To validate our core hypothesis that joint training enables a mutually beneficial information exchange capable of driving robust zero-shot real-world generalization for both our objectives, it is important that the shared U-Net $f_\theta$ effectively handles these modalities and domain disparity.

Taking inspiration from prior literature \cite{geowizard, wonder3d}, we initially evaluated a strategy that makes use of a single U-Net $f_\theta$ to handle multi-modal data via a task switcher, denoted as $s$. The switcher $s$, a one-dimensional vector that labels different tasks, $s \in \{s_d, s_s\}$ (depth estimation and segmentation, respectively), is mapped through a low-dimensional positional encoder and added directly to the network's time embeddings. Formulated as:

\begin{equation}
\label{eq:old_switcher}
\hat{{z}}^{d} = f_\theta({z}^{{x}_d}, t, s_d), \quad \hat{{z}}^{s} = f_\theta({z}^{{x}_s}, t, s_s)
\end{equation}

However, as detailed in Sec. \ref{sec:ablation}, this configuration quickly collapses. Hence, to establish a more robust task-routing mechanism, we re-examine the core architecture of our generative prior. Intuitively, for a text-to-image model such as Stable Diffusion, which was trained on internet-scale datasets to perform high-resolution image generation explicitly conditioned on text prompts encoded by CLIP~\cite{clip}, it must maintain a profound structural connection between its language prompts and its visual outputs. However, previous monocular depth estimators have largely ignored this prompt guidance pathway \cite{marigold, lotus, wonder3d, geowizard}, feeding empty text prompts to extract only visual features. In contrast, we propose a fine-tuning protocol to fully exploit this pre-trained text-image alignment. 
Since a visual synthesis model inherently possesses a deep structural understanding of scenes, leveraging CLIP's shared embedding space for task conditioning provides a significantly more effective mechanism for routing cross-modal information.

To provide robust global guidance, we explicitly condition each pathway using task-specific text prompts: $y_s = \text{``predict glass segmentation"}$ and $y_d = \text{``predict depth map"}$. Using a frozen CLIP text encoder $\tau_\theta$, we extract the corresponding text embeddings $c_s = \tau_\theta(y_s)$ and $c_d = \tau_\theta(y_d)$. Rather than only training a naive positional vector from scratch to handle task switching, we inject these embeddings directly into the U-Net via its pre-trained cross-attention layers, as in Stable Diffusion. This approach seamlessly maps our explicit task instructions into the network's deep feature space, fully leveraging the rich, pre-trained priors of Stable Diffusion U-Net, which was trained on text conditioning. Consequently, our final joint fine-tuning loss function is defined as:
\begin{equation}
\begin{split}
L_T =\;& 
\| {z}^{s} - f_\theta({z}^{{x}_s}, t, s_s, c_s,) \|_2^2\\&+ \| {z}^{d} - f_\theta({z}^{{x}_d}, t, s_d, c_d,) \|_2^2
\end{split}
\label{eq:our_loss_function}
\end{equation}
\noindent
As demonstrated in our experimental results (Sec. \ref{sec:results}), our pipeline shows impressive zero-shot transfer, beating state-of-the-art in both glass segmentation and glass-aware depth estimation.

\subsection{Inference and Recovering Metric Depth to Glass}
\label{sec:glass_metric_depth}
\subsubsection{Inference}

During inference, the input RGB image is first processed by the frozen Encoder $\mathcal{E}$ to extract the image latent ${z}^{x}$. That is then passed through the shared U-Net alongside the task-specific CLIP text embeddings ($c_s$ and $c_d$) which route the network's features, enabling it to predict the glass segmentation latent $\hat{{z}}^{s}$ and the glass-aware depth latent $\hat{{z}}^{d}$ in a single forward pass without task interference.  Both predicted latents are subsequently projected back into pixel space via the frozen Decoder $\mathcal{D}$. To recover the final single-channel predictions for both modalities, we average the three decoded output channels. Finally, a fixed threshold is applied to the segmentation output to produce the definitive binary glass mask.

\subsubsection{Metric Depth for Glass Surfaces}
The raw sensor depth $D_{raw}$ from an RGB-D camera is converted to raw disparity $D'_{raw} = 1 / D_{raw}$. Subsequently, the model-predicted affine-invariant disparity $\hat{d}$ and a background mask $M_{bg}$ (which excludes the model-predicted glass regions) are used to recover the global scale $s$ and shift $t$. This is achieved by solving a linear least-squares optimization over the valid depth points in the background pixels:
\begin{equation}
\arg\min_{s, t} \sum_{i \in M_{bg}} (s \cdot \hat{d}_{i} + t - D'_{raw, i})^2
\end{equation}

Once $s$ and $t$ are computed, we apply this affine transformation to the entire predicted disparity map and invert it to recover the aligned metric depth $D_{aligned} = 1 / (s \cdot \hat{d} + t)$

Finally, we construct a fused, glass-aware depth map by retaining the reliable raw sensor depth for the background and injecting our aligned metric predictions within the predicted glass regions. This compositing strategy completely resolves sensor failure artifacts on transparent glass surfaces, yielding an accurate glass-aware depth map. The depth map is then converted to a point cloud using known camera intrinsics.

\section{EVALUATION}
To train and evaluate SILICA for joint glass segmentation and monocular depth estimation, we use separate data sources for each task while maintaining balanced multi-task supervision. All datasets and evaluation protocols are organized under two core categories: Glass Segmentation and Monocular Depth Estimation. Training our method to convergence takes approximately 1 day on a single Nvidia RTX 5090 GPU card.

\subsection{Mirage 18k}

\label{subsec:mirage_18k}

We introduce the Mirage 18k Dataset, comprising 18,353 manually annotated images with paired binary segmentation masks, spanning 38 unique scenes. We hand-picked 12,420 samples strictly for segmentation training and 3,527 strictly for segmentation testing. To ensure there is no overlap between the two sets, we make sure that no data taken from one scene is part of both the test set and the training set. The remaining 2,406 samples include paired ground-truth metric depth to glass in addition to the binary segmentation masks, intended strictly for zero-shot depth and segmentation evaluation.

\begin{figure}[htbp]
    \centering
    \includegraphics[width=.8\columnwidth]{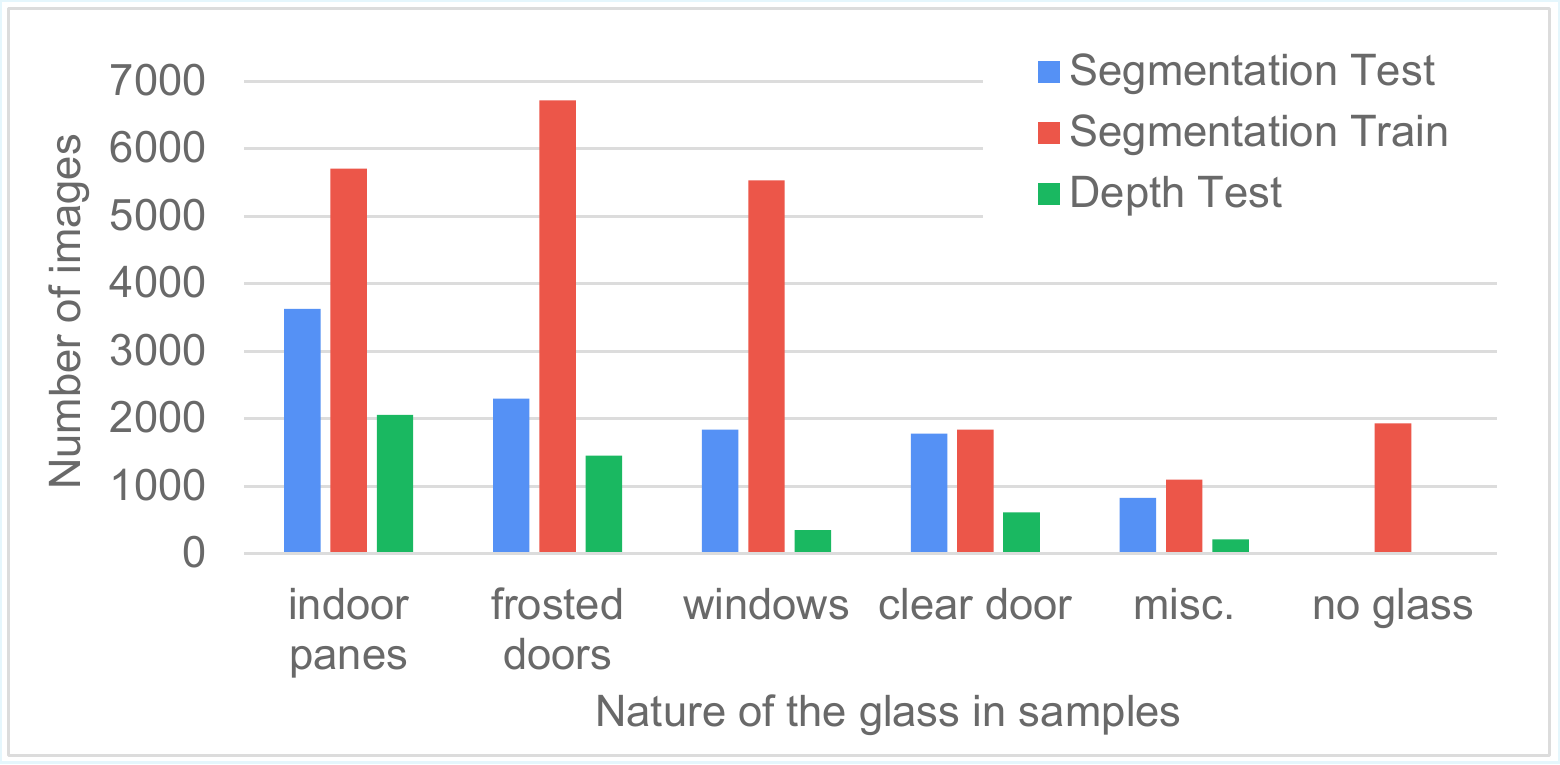}
    \caption{\textbf{Diversity of type of glass in Mirage 18k.} Mirage 18k provides training and testing data for several types of glass, often with multiple in the same image. 1,930 samples without any glass are also included to provide negative samples in the training set.}
    \label{fig:images-bar}
    \vspace{-0.2cm}
\end{figure}

Our dataset captures a diverse distribution of glass structures in the wild: we have complex scenes in which multiple types of glass structures (e.g., a frosted door next to an indoor pane) are in the same image. Fig.~\ref{fig:images-bar} shows the distribution quantitatively.

\begin{figure}[h]
    \centering
    \includegraphics[width=0.75\linewidth]{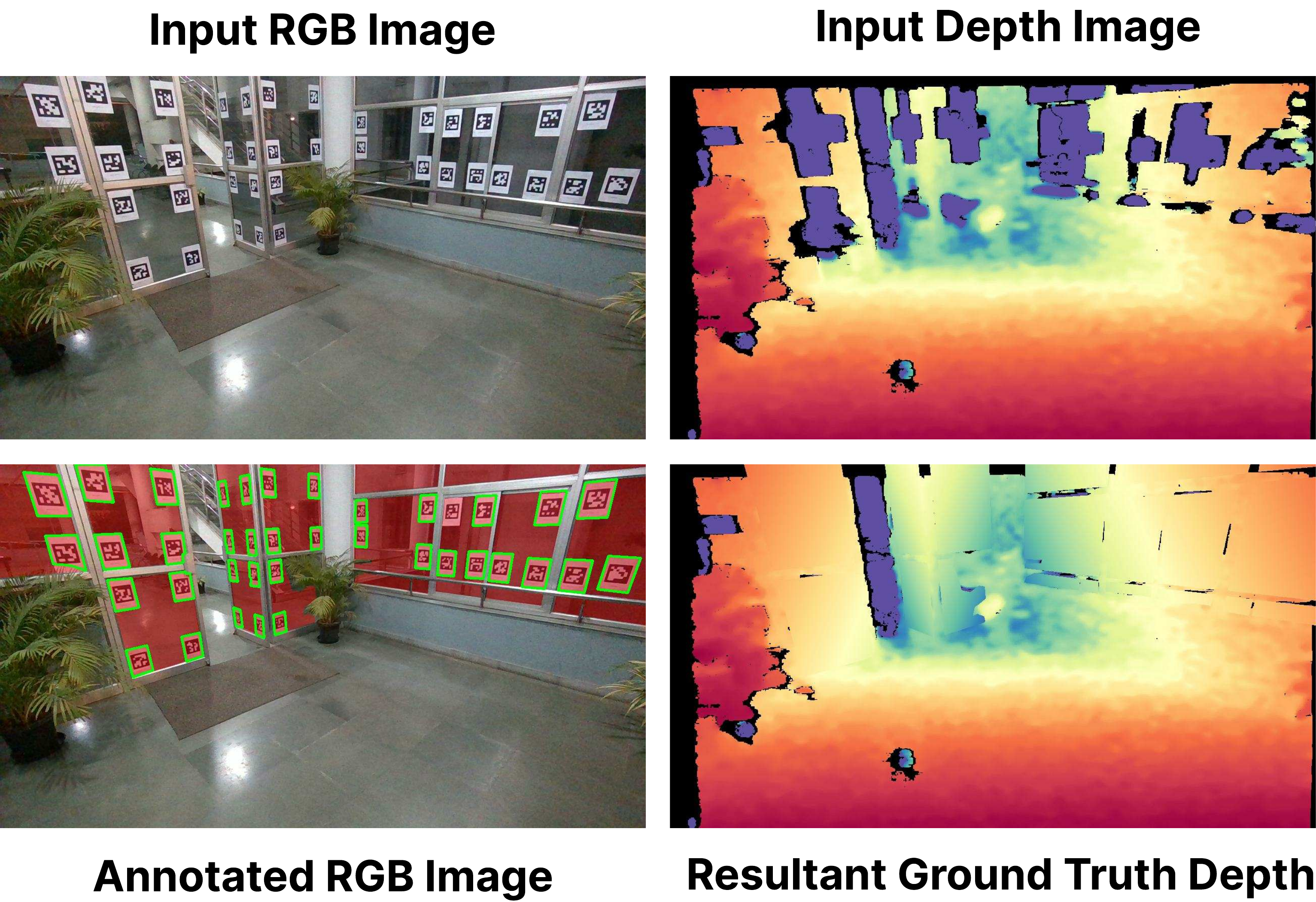}
    \caption{\textbf{Acquisition of ground truth depth for Mirage 18k.} The raw sensor depth image gives noisy returns for glass areas, as it is transparent to IR. The corrected depth image shows polygons with smoothly varying depth to glass, which is used for validating our results.}
    \label{fig:gt-demo}
    \vspace{-5mm}
\end{figure}

We use the CVAT annotation tool to generate ground truth binary glass segmentation masks. Ground truth depth to glass is obtained by modeling each pane as a 3D planar surface. Temporary opaque paper markers were placed at the corners of each pane and annotated for in CVAT, and their median depth was extracted from a local $15 \times 15$ pixel neighborhood. To ensure the frames line up perfectly, we record data twice: a completely stationary robot first records the glass adorned with the markers to capture reliable metric depth to glass surfaces, and then records a second sequence from the exact same pose after the markers are removed. By solving an overdetermined system via least-squares optimization in the inverse depth space, we reconstruct dense metric depth maps, as illustrated in Fig.~\ref{fig:gt-demo}. This ground truth depth is used strictly for depth evaluation, and not for training.

\subsection{Training Datasets}

\subsubsection{Glass Segmentation}
We aggregate the training splits of our custom Mirage 18k dataset together with several established binary glass segmentation datasets: 3DRef~\cite{3DRef}, Trans10k~\cite{TransLab_Trans10k}, GDD~\cite{GDD_DGNet}, GSD-S~\cite{GlassSemNet_GSDS}, and GW-Depth~\cite{GWDepth}. This curation yields a total of 24,811 training samples, covering diverse indoor environments with planar and curved transparent structures, varying lighting conditions, and cluttered backgrounds.

\subsubsection{Monocular Depth Estimation} For depth estimation, we utilize Hypersim~\cite{hypersim}, a large-scale, photorealistic synthetic dataset providing high-fidelity depth annotations across 461 indoor scenes. To maintain a balanced multi-task learning objective and prevent task imbalance during training, we filter out incomplete data and randomly sample exactly 24,811 images from the official training split to perfectly match the size of our glass segmentation corpus.

\subsection{Evaluation Datasets}
\subsubsection{Glass Segmentation} We evaluate segmentation performance on the official test splits of GDD~\cite{GDD_DGNet}, 3DRef~\cite{3DRef}, Trans10k~\cite{TransLab_Trans10k}, and our Mirage 18k dataset. To evaluate pure out-of-distribution generalization, we also test on a completely unseen, in-the-wild our shopping mall dataset.

\subsubsection{Monocular Depth Estimation} We evaluate our depth estimation performance across three real-world datasets. Two of these are domain-specific datasets (Mirage 18k and GW-Depth) that provide explicit depth annotations for transparent glass structures. The third is NYUv2~\cite{nyuv2}, a widely adopted general indoor depth benchmark, which we use to demonstrate the model's robust generalization in standard environments.

\subsection{Baselines and Evaluation Protocols}

\subsubsection{Glass Segmentation}
We report the mean Intersection-over-Union (mIoU) using the official TorchMetrics implementation~\cite{torch_metrics}. To ensure a fair comparison against prior methods, we apply the following protocols:
\begin{itemize}
    \item TransLab and EBLNet are trained and evaluated on their official splits. Due to the high computational requirements of EBLNet, we report results that represent the most feasible training configuration.
    \item GDNet results are reported strictly on datasets for which pretrained models are publicly available.
    \item GlassSemNet requires full-scene semantic annotations. In cases where retraining was not feasible, we report inference-only results.
\end{itemize}

\subsubsection{Monocular Depth Estimation}
We compare our monocular depth estimation pathway against one domain-specific architecture, GW-Depth~\cite{GWDepth} (which explicitly handles glass using a dual-context transformer to exploit boundary and reflection cues), as well as two state-of-the-art foundation models for discriminative monocular depth estimation models: Lotus-D~\cite{lotus} and Depth Anything V2~\cite{depth_anything_v2}.

\textit{Evaluation Protocol \& Metrics:} For all depth evaluations across the three datasets (Mirage 18k, GW-Depth, and NYUv2), we follow the standard affine-invariant depth estimation protocol utilized by Marigold~\cite{marigold}. This protocol aligns the estimated depth predictions with the available ground truth using standard least-squares fitting across all valid pixels to recover the global scale and shift. 

The accuracy of these aligned predictions is then assessed using standard monocular depth metrics. We report the absolute mean relative error (AbsRel) $\frac{1}{M} \sum_{i=1}^{M} |a_i - d_i|/d_i$ where $M$ is the total number of evaluated pixels, $a_i$ is the aligned predicted depth map, and $d_i$ represents the ground truth depth. We also report threshold accuracies $\delta_1$ and $\delta_2$, which denote the proportion of pixels satisfying $\max(a_i/d_i, d_i/a_i) < 1.25$ and $< 1.25^2$, respectively.

\section{RESULTS}
\label{sec:results}
Our qualitative and quantitative results demonstrate the successful integration of Stable Diffusion priors with CLIP-conditioned task switching. That enables highly accurate and generalized zero-shot joint prediction of glass segmentation and glass-aware monocular depth. This proves our initial intuition, i.e., by jointly training both modalities, the mutual information exchange allows the network to establish a strict foreground-background hierarchy. Furthermore, our ability paints the importance of CLIP conditioning and empirically prove that paired glass depth annotation is not required.

\begin{table}[htbp]
    \centering
    \caption{Segmentation mean Intersection over Union (mIoU) Results}
    \label{tab:segmentation_iou}
    \setlength{\tabcolsep}{3pt}
    \begin{tabular*}{\columnwidth}{@{\extracolsep{\fill}}l ccccc}

        \toprule
        \multirow{2}{*}{\textbf{Method}} & \multicolumn{5}{c}{\textbf{Datasets}} \\
        \cmidrule(lr){2-6}
        & \textbf{GDD} & \textbf{3DRef} & \textbf{Trans10k} & \textbf{Mirage 18k} & \textbf{Mall} \\
        \midrule

        TransLab~\cite{TransLab_Trans10k} & 88.07 & 83.06 & 78.05 & 72.69 & 59.09 \\
        EBLNet~\cite{eblnet} & 88.72 & 86.71 & 90.28 & 51.05 & 44.39 \\
        GDNet\textsuperscript{\textdagger}~\cite{GDD_DGNet} & 87.64\textsuperscript{*} & 84.46 & 77.25 & 67.20 & 42.06 \\
        GlassSemNet\textsuperscript{\textdagger}~\cite{GlassSemNet_GSDS} & 90.80\textsuperscript{*} & 53.13 & 78.58 & 77.50 & 65.55 \\
        SAM3\textsuperscript{\textdagger}~\cite{sam3} & 65.69 & 74.37 & 67.93 & 77.51 & 61.03 \\

        \textbf{SILICA (Ours)} & \textbf{91.60} & \textbf{88.31} & \textbf{92.43} & \textbf{81.21} & \textbf{77.83} \\

        \bottomrule
    \end{tabular*}
    
    \vspace{0.15cm}
    
    \begin{minipage}{\columnwidth}
    \footnotesize{\textsuperscript{\textdagger} These models were evaluated using their provided pre-trained weights and were not (re)trained by us for these experiments.} \\
    \footnotesize{\textsuperscript{*} metrics were taken from respective literature}
    \end{minipage}
    \vspace{-0.4cm}
    
\end{table}

\subsection{Quantitative Results}

\subsubsection{Glass Segmentation} 

As shown in Tab.~\ref{tab:segmentation_iou}, SILICA consistently outperforms prior domain-specific baselines across all evaluated public datasets. To assess zero-shot generalization in complex, real-world environments, we further evaluate all models on an unobserved, internal shopping mall deployment dataset featuring large-scale glass architectures. Notably, our approach demonstrates superior performance, surpassing all domain-specific baselines but also the recent state-of-the-art foundational segmentation model, SAM3.

\subsubsection{Monocular Depth Estimation}

As reported in Tab.~\ref{tab:depth_metrics}, our joint prediction pipeline consistently outperforms the domain-specific baseline, GW-Depth, across all evaluated datasets, even though GW-Depth model was explicitly trained on both GW-Depth dataset and NYUv2 dataset we still outperform them. Furthermore, when compared against state-of-the-art discriminative monocular depth foundation models, our approach yields the highest performance on datasets featuring prominent transparent structures, achieving improvements of up to 20\% in specific metrics. 

Crucially, standard depth sensors fail on transparent surfaces, frequently capturing background objects instead of the physical glass. We explicitly evaluate this using our Mirage 18k depth test-only set. As shown in Tab.~\ref{tab:glass_only}, failing to segment and filter these invalid sensor points prior to metric alignment causes a catastrophic error spike of over 85\% compared to our filtered strategy. This demonstrates that our joint prediction framework is not merely an algorithmic improvement, but an important prerequisite for reliable 3D mapping and safe, collision-free autonomous navigation in real-world robotic deployments.

\begin{table*}[t]
    \centering
    \caption{Quantitative Results for Depth Estimation}
    \label{tab:depth_metrics}

    \setlength{\tabcolsep}{5pt}
    \renewcommand{\arraystretch}{1.15}

    \begin{tabular}{l c ccc ccc ccc}
        \toprule
        \multirow{2}{*}{\textbf{Method}} 
        & \multirow{2}{*}{\textbf{Task}}
        & \multicolumn{3}{c}{\textbf{NYUv2}~\cite{nyuv2}} 
        & \multicolumn{3}{c}{\textbf{GW-Depth}~\cite{GWDepth}}
        & \multicolumn{3}{c}{\textbf{Mirage 18k (Ours)}}  \\
        \cmidrule(lr){3-5} 
        \cmidrule(lr){6-8}
        \cmidrule(l){9-11}
        & 
        & \textbf{AbsRel} $\downarrow$ 
        & \textbf{$\delta_1$} $\uparrow$ 
        & \textbf{$\delta_2$} $\uparrow$ 
        & \textbf{AbsRel} $\downarrow$ 
        & \textbf{$\delta_1$} $\uparrow$ 
        & \textbf{$\delta_2$} $\uparrow$
        & \textbf{AbsRel} $\downarrow$ 
        & \textbf{$\delta_1$} $\uparrow$ 
        & \textbf{$\delta_2$} $\uparrow$ \\
        \midrule

        Lotus-D~\cite{lotus}
        & Depth
        & 5.1 & 97.2 & 99.2 & 6.4 & 95.6 & 99.11 & 19.24 & 79.88 & 92.99 \\

        Depth Anything V2~\cite{depth_anything_v2}
        & Depth
        & \textbf{4.5} & \textbf{97.9} & \textbf{99.3} & 5.08 & 97.37 & \textbf{99.7} & 25.78 & 68.60 & 85.76 \\

        GW-Depth~\cite{GWDepth}
        & Depth + Seg
        & 9.8$^{\dagger}$ & 91.17$^{\dagger}$ & 99.0$^{\dagger}$ & 10.0$^{\dagger}$ & 90.0$^{\dagger}$ & 98.9$^{\dagger}$ & 24.39 & 49.11 & 78.86 \\

        
        \textbf{SILICA (Ours)}
        & Depth + Seg
        & 5.3 & 97.1 & \textbf{99.3} & \textbf{4.4} & \textbf{98.14} & \textbf{99.7} & \textbf{18.04} & \textbf{95.34} & \textbf{97.57} \\

        \bottomrule
    \end{tabular}

    \vspace{0.2cm}
    \parbox{0.95\textwidth}{\footnotesize
    $^{\dagger}$ GW-Depth was trained specifically on NYUv2.
    }
\end{table*}

\begin{table}[t]
    \centering
    \caption{Depth Estimation Performance on Glass Regions after aligning using raw sensor points}
    \label{tab:glass_only}
    \begin{tabular}{l c ccc}
        \toprule
        \textbf{Model} 
        & \textbf{Task}
        & \textbf{AbsRel} $\downarrow$
        & \textbf{$\delta_1$} $\uparrow$
        & \textbf{$\delta_2$} $\uparrow$ \\
        \midrule

        Lotus-D$^{\dagger}$~\cite{lotus}
        & Depth
        & 161.60 & 2.80 & 73.40 \\

        Depth Anything V2$^{\dagger}$~\cite{depth_anything_v2}
        & Depth
        & 174.90 & 5.60 & 56.20 \\

        GW-Depth~\cite{GWDepth}
        & Depth + Seg
        & 32.90 & 27.50 & 58.40 \\

        \textbf{SILICA (Ours)}
        & Depth + Seg
        & \textbf{24.73} & \textbf{94.30} & \textbf{97.72} \\
        \bottomrule
    \end{tabular}

    \vspace{0.2cm}
    \parbox{0.95\textwidth}{\footnotesize
    $^{\dagger}$ Glass mask was not applied to remove the wrong depth.
    }
    \vspace{-0.7cm}
\end{table}

\begin{figure*}
    \centering
    \includegraphics[width=1.0\linewidth]{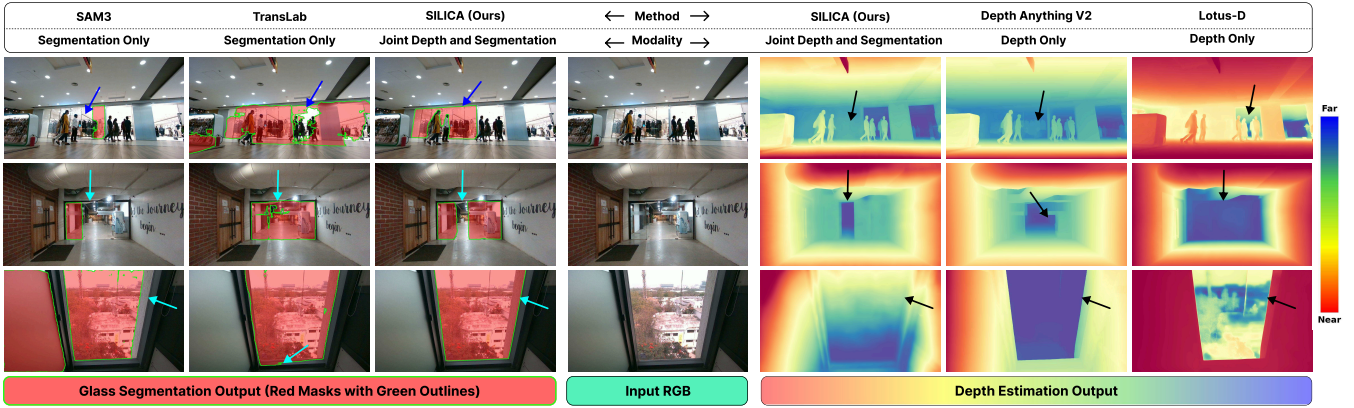}
    \caption{\textbf{Qualitative comparison of glass segmentation and depth prediction}. SILICA can jointly predict glass segmentation and glass-aware depth estimation precisely even in scenes with a busy background (seen in row 1), a half-open glass door (seen in row 2), and a glass window (seen in row 3). SILICA shows consistent performance in all these samples, even in the midst of saliency. The joint prediction approach also helps it correctly identify and predict depth to challenging open/closed doors where other methods hallucinate flat glass or miss the door (row 2).}
    \label{fig:qualitative}
    \vspace{-0.5cm}
\end{figure*}

\subsection{Qualitative Results}
As presented in Fig.~\ref{fig:qualitative}, we provide a qualitative comparison between our approach and the top-performing baselines across highly challenging environments typical of indoor robotic deployment. Notably, these complex, real-world scenarios are uniquely introduced by our novel Mirage 18k dataset to explicitly benchmark operational robustness.

\textbf{\ding{172} Crowded shopping mall (large glass panes):} Our model successfully ignores dynamic obstacles (humans) and accurately segments the glass despite complex background details. In contrast, TransLab and SAM3 fail due to background and foreground saliency. For depth estimation, while DepthAnythingv2 incorrectly captures the depth of objects behind the glass, our model accurately predicts the glass depth, while Lotus-D also gave decent predictions but fails in consistency.

\textbf{\ding{173} Half-opened glass door:} TransLab fails to predict the open door, and SAM3 misses the door structure entirely, segmenting only the isolated pane. Our model, however, extracts a highly accurate segmentation mask. In terms of depth, Lotus-D wrongly predicts a fully closed scene at an incorrect distance, and DepthAnythingV2 struggles significantly on the right side due to background saliency. In contrast, out model predicts accurate surface depth.

\textbf{\ding{174} Extreme close-up window (minimal context):} Given highly constrained viewpoints, TransLab and SAM3 predict incorrect boundaries influenced by a background pillar. Our model precisely isolates the true, closer glass boundary. Furthermore, Lotus-D and DepthAnythingV2 completely miss the transparent surface, outputting background depth, while our model consistently resolves the correct glass depth.

These qualitative results provide clear evidence that our joint formulation effectively learns a robust spatial hierarchy, successfully resolving severe visual ambiguities in the presence of complex transparent structures.

\subsection{Real-world Deployment}
\label{sec:deployment}
Designed explicitly for robotics, SILICA is integrated into our custom autonomous wheelchair stack for real-world navigation. To operate in a new environment, we first collect a map by feeding the corrected glass depth points from our pipeline into RTAB-Map \cite{labbe2019rtab} to construct an accurate 3D map. From which, we obtain a 2D occupancy grid map where the glass areas are included as obstacles. These maps are then utilized for our MPPI-based 2D navigation and 3D localization, enabling the wheelchair to safely plan paths and navigate through glass-structured environments (as shown in Fig.~\ref{fig:fig_teaser}). Crucially, multiple deployment runs verify that our model produces very rare false positives in glass-free scenes, ensuring highly reliable robotic operation. Running on a laptop with an NVIDIA RTX 3060 GPU, the pipeline achieves an inference rate of approximately 5~Hz for joint segmentation and depth estimation. Because glass structures are static, we directly inject these predictions into a spatio-temporal voxel grid with a decay mechanism to ensure smooth real-time performance. Finally, we also observe that our pipeline implicitly improves overall 3D mapping, as we provide RTAB-Map with stabilized depth points on glass surfaces results in significantly better and cleaner final maps.

\begin{figure}[h]
    \centering
    \includegraphics[width=\linewidth]{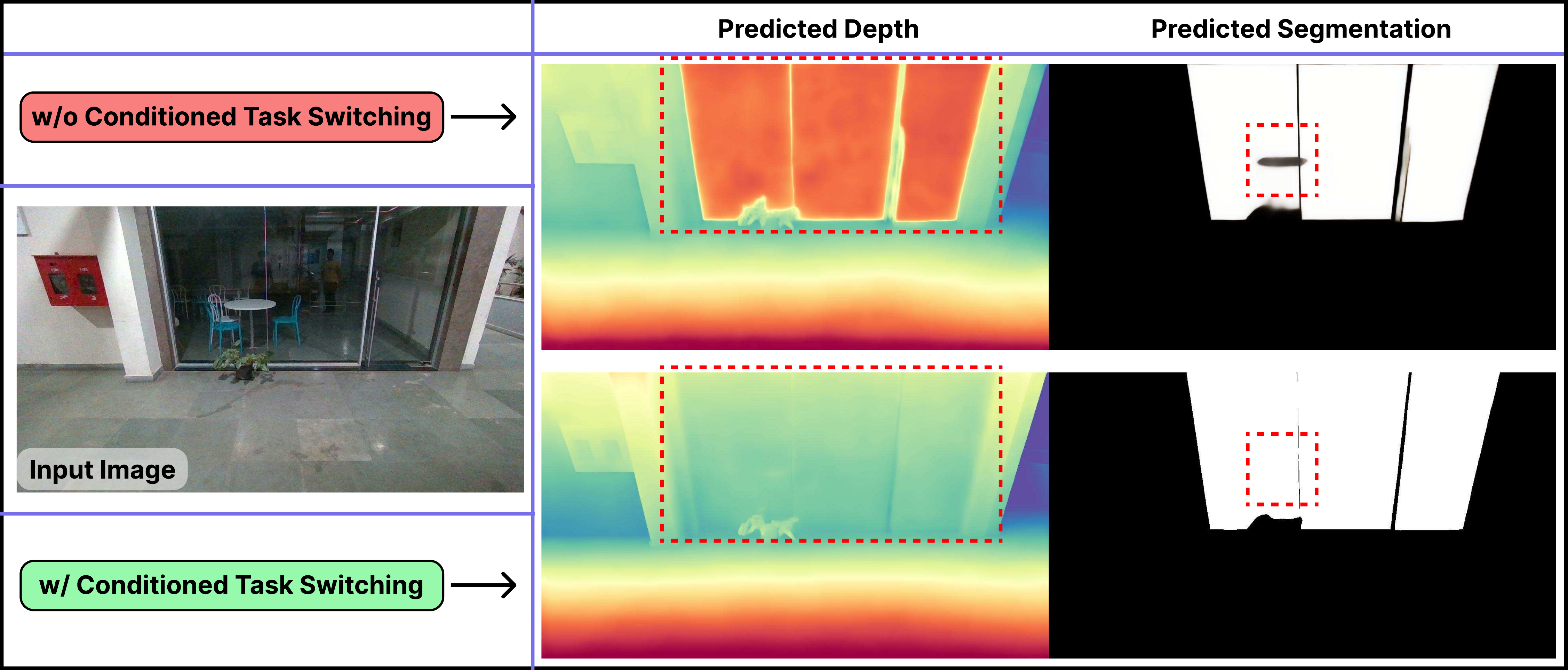}
    \caption{(Top) Without conditioning, task entanglement corrupts depth predictions and segmentation. (Bottom) SILICA's CLIP-conditioned cross-attention enables stable mutual information exchange, successfully resolving the foreground-background hierarchy.}
    \label{fig:abilation_confusion}
    \vspace{-4mm}
\end{figure}

\subsection{Ablation}
\label{sec:ablation}

\textbf{CLIP-Conditioning:} To justify our architectural design, we ablate our CLIP-based conditioning module against a naive positional encoder for task switching (Eq.~\ref{eq:old_switcher}). As established earlier, the severe scarcity of real-world glass depth annotations necessitates training on a complex mixture of synthetic depth and real-world segmentation data. Under these multi-modal conditions, simple positional encoding suffers from severe task interference. As shown in Fig.~\ref{fig:abilation_confusion}, the baseline fails to establish a proper spatial hierarchy: it wrongly predicts the glass surface as abnormally close in the depth map and also fails foreground-background disambiguation in the segmentation mask (e.g., misclassifies background furniture as foreground).

In contrast, our CLIP-conditioned approach (detailed in Sec.~\ref{sec:clip_c_joint_prediction}) connects directly to all cross-attention layers of the U-Net, fully leveraging the robust priors of the diffusion model. This mechanism enables highly stable task switching and facilitates effective mutual information transfer between the depth and segmentation pathways. As seen in Fig.~\ref{fig:abilation_confusion}, this formulation successfully enforces a strict spatial hierarchy, resolving the visual ambiguity and dramatically improving the output quality of both tasks. This empirically proves that CLIP conditioning is essential to bridge the sim-to-real domain gap and successfully execute our joint-prediction framework.

\textbf{Eliminating glass depth annotations:} To test if explicit glass supervision is necessary, we filtered Hypersim, discarding scenes with $>10\%$ glass pixels. Retraining from scratch on 24,000 strictly glass-free images yielded only marginal metric differences (Tab.~\ref{tab:ablation_glass}). This robust performance definitively validates our core intuition: explicit glass depth annotations are not strictly required. Instead, our CLIP-conditioned joint training framework teaches the network the correct spatial hierarchy for transparent surface perception.

\textbf{Mutual Benefit of Joint Training:} Ablating SILICA against single-task baselines proves their strict interdependence. A segmentation-only model suffers a $\sim 5.0\%$ mIoU drop with increased real-world false positives. Conversely, a depth-only model lacks the predicted segmentation mask essential for explicitly filtering corrupted raw sensor data during metric alignment, resulting in a $>80\%$ error spike on glass regions.

\begin{table}[t]
    \centering
    \caption{Effect of Glass Scenes in Hypersim on SILICA Model performance.}
    \label{tab:ablation_glass}
    \begin{tabular}{l ccc}
        \toprule
        \textbf{Variant} 
        & \textbf{AbsRel} $\downarrow$ 
        & \textbf{$\delta_1$} $\uparrow$ 
        & \textbf{$\delta_2$} $\uparrow$ \\
        \midrule
        
        \multirow{1}{*}{w/o Glass Scenes}
        & 18.93 & 94.47 & \textbf{97.68} \\

        
        \multirow{1}{*}{w/ Glass Scenes}
        & \textbf{18.04} & \textbf{95.34} & 97.57 \\

        \bottomrule
    \end{tabular}
    \vspace{-0.3cm}
\end{table}

\section*{ACKNOWLEDGMENT}
The authors used large language models (Gemini, ChatGPT) and Grammarly strictly to improve the manuscript's linguistic flow. We also thank IHub-Data for their support via project M2-029.

\printbibliography
\addtolength{\textheight}{-12cm}

\end{document}